\documentclass{article}

\usepackage{PRIMEarxiv}

\usepackage{caption}
\usepackage{subcaption}
\usepackage{tikz}
\usepackage[bookmarks=false]{hyperref}
    \hypersetup{colorlinks,
      linkcolor=blue,
      citecolor=blue,
      urlcolor=blue}
\newtheorem{theorem}{Proposition}
\usepackage[figuresright]{rotating}
\usepackage{algorithm}
\usepackage{algpseudocode}
\usepackage{bm}
\newcommand{\ba}{\mathbf{a}}
\newcommand{\bs}{\mathbf{s}}
\newcommand{\bt}{\mathbf{t}}
\newcommand{\bx}{\mathbf{x}}
\newcommand{\by}{\mathbf{y}}

\usepackage[utf8]{inputenc} 
\usepackage[T1]{fontenc}    
\usepackage{hyperref}       
\usepackage{url}            
\usepackage{booktabs}       
\usepackage{amsfonts}       
\usepackage{nicefrac}       
\usepackage{microtype}      
\usepackage{lipsum}
\usepackage{fancyhdr}       
\usepackage{graphicx}       
\graphicspath{{media/}}     

\title{Locally Smoothed Gaussian Process Regression
}

\author{
  Davit Gogolashvili, \quad Bogdan Kozyrskiy \quad Maurizio Filippone\\
  \\
  Department of Data Science, EURECOM \\
  450 Route des Chappes, 06410 Biot, France\\
  \texttt{\{davit.gogolashvili, bogdan.kozyrskiy, maurizio.filippone\}@eurecom.fr} \\
}

\begin{document}
\maketitle

\begin{abstract}
We develop a novel framework to accelerate Gaussian process regression (GPR).
In particular, we consider localization kernels at each data point to down-weigh the contributions from other data points that are far away, and we derive the GPR model stemming from the application of such localization operation.
Through a set of experiments, we demonstrate the competitive performance of the proposed approach compared to full GPR, other localized models, and deep Gaussian processes.
Crucially, these performances are obtained with considerable speedups compared to standard global GPR due to the sparsification effect of the Gram matrix induced by the localization operation. 
\end{abstract}


\section{Introduction}
Function estimation is a fundamental problem in Machine Learning. 
In supervised learning tasks applied to a data set composed of observed input data and labels, the goal of function estimation is to establish a mapping between these two groups of observed quantities. 
Function estimation can be approached in various ways, and we can broadly divide algorithms in two categories, as \textit{global} and \textit{local}. 
Examples of global algorithms are Neural Networks \cite{Neal96} and kernel machines \cite{Shawe-Taylor04}, which impose a functional form yielding a global representation of the function. The functional form is parameterized by a set of parameters which are optimized or inferred based on all the available data. The estimated model can later be used to query the function at any input points of interest. In local algorithms such as K-Nearest Neighbors (KNN), instead, the target point is fixed and the corresponding value of the function is estimated based on the closest data available.

Obviously, any global algorithm can be made local by training it only for the few training points located in the vicinity of the target test point. While it may seem that the idea of localizing global algorithms is not a very profound one, empirical evidence shows that localization could improve the performance of the best global models \cite{bottou1992local}. The idea of localization was therefore applied to global models such as SVMs \cite{blanzieri2006adaptive, blanzieri2008nearest}. In addition to performance gains, by operating on smaller sets of data points, these local approaches enjoy computational advantages, which are particularly attractive for kernel machines for which scalability with the number of data points is generally an issue \cite{cheng2007localized,segata2010fast, segata2012local}.

In this work, we develop novel ideas to implement a localization of Gaussian processes (GPs) in order to obtain performance gains, as well as computational ones.
GPs are great candidates to benefit from computational speedups given that a naïve implementation requires expensive algebraic computations with the covariance matrix; denoting by $n$ the number of input data, such operations cost $\mathcal{O}(n^3)$ operations and require storing $\mathcal{O}(n^2)$ elements, hindering the applicability of GPs to data sets beyond a few thousand data points \cite{quinonero2005unifying}. 
Another issue with GPs is how to choose a suitable kernel for the problem at hand so as to avoid problems of model misspecification.  
Both of these issues have been addressed in various ways, by proposing scalable approximations based on inducing points \cite{HensmanGPsForBigData} and random features \cite{Rahimi08,Cutajar17}, and by composing GPs to obtain a rich and flexible class of equivalent kernels \cite{Wilson16b}.


In this work we explore an alternative way to address scalability and kernel design issues by localizing GPs. 
In particular, we show how the localization operation leads to a particular form for the localized GP, and what is the effect on the kernel of this model. 
Furthermore, the localization makes it apparent how to implement the model with considerable gains compared to other approaches to approximate GPs.
We demonstrate such performance gains on regression tasks on standard UCI benchmarks \cite{Asuncion07}.

\subsection{Related work}
    {\em Local learning algorithms}  were introduced by Bottou and Vapnik \cite{bottou1992local}, with the main objective of estimating the optimal decision function for each single testing point. Examples of local learning algorithms include the well-known K-Nearest Neighbor regression \cite{altman1992introduction}  and local polynomial regression \cite{fan2018local}. These methods provide simple means for solving regression problems for the cases where training data are nonstationary or their size is prohibitively large for building a global model. 
    However, neither of these methods provides ways to quantify uncertainty in predictions, which is a highly desirable feature in cost-sensitive applications. 
    
    Gaussian Process Regression  (GPR) \cite{Rasmussen06} is a popular nonparametric regression method based on Bayesian principles which provides uncertainty estimates for its predictions. Similarly to other kernel methods (e.g., SVMs and KRR), GPR is a global method, meaning that it takes into account the whole dataset at prediction time. Thus, GPR inherits the computational complexity of global kernel methods, which is prohibitive for large datasets. Among the large class of scalable approximations for GPR, successful ones are based on Random Fourier Features \cite{Rahimi08} and on sparsification of the Gram matrix induced by the kernel \cite{Rasmussen06}.
    
    Random feature approximation of the kernel proposed in \cite{Rahimi08} is based on Bochner theorem and allows representing  the kernel function as a dot product of (possibly infinite) feature maps applied to the input data.
    In practice, infinite feature maps are replaced by a finite Monte Carlo approximation.
    The disadvantage of this approach is that it is necessary to construct a specific random feature mapping for each type of kernel. 
    While random feature approximations are known for popular kernels such as RBF \cite{Rahimi08} and polynomial \cite{NIPS2015_f7f580e1}, there is no straightforward application of this method to approximate arbitrary kernels.
    
    The Gram matrix sparsification approach is based on the idea of introducing so-called inducing points in order to approximate the full Gram matrix. 
    One of the most popular methods in this family is the Nyström approximation \cite{Rasmussen06}.
    The main drawback of this approach is that a low number of inducing points might lead to a poor approximation of the original model, which affects predictive performance.    
    An important advancement within this family of approaches which provides a scalable variational formulation was proposed in \cite{Titsias09}. 

    While providing good performance and scalability for large datasets, these approaches still require some design choices for the kernel. 
    For stationary kernels, they assume that the same kernel is suitable for all the regions of input space and if data are nonstationary, this may harm the predictive performance. 
    The literature has a wide range of proposals to address kernel design by incorporating ideas from deep learning \cite{Wilson16b,cutajar2017random}. 
    
    Recently, partitioning strategies have also gained some attention. The main idea is to divide the input space in regions where local estimators are defined \cite{meister2016optimal,tandon2016kernel,muecke2019reducing,carratino2021park}. In partition-based methods, the main challenge is to define an effective partitioning of the space.
    
    There are several approaches that use the idea of local learning for training GP models. The method proposed in \cite{meier2014local} and extended  in \cite{meier2014incremental} mostly focuses on Bayesian parametric linear regression. The methods in these papers build an ensemble of local models centered at several fixed points, where each training point is weighted accordingly to the distance from the center of the model. Predictions are computed as a weighted sum of the local models. The authors claim that their approach extends to GPR, but in this case each local model considers the full training set. This means that these methods use localization to address nonstationarity, but poorly scale to large datasets. 
    The method proposed in \cite{snelson2007local} proposes to build local GP models that use only subsets of the training data,  but it lacks a mechanism that assigns importance weight for the training points for each local model according to the distance from the center of the model. That is why the model can make overconfident predictions for the points that lay far away from the centers of the local models.    In \cite{gramacy2015local}  in order to obtain fast approximate prediction at a target point, the Authors propose a forward step-wise variable selection procedure to find the optimal sub-design.

The paper is organized as follows:
Section 2 introduces GPs and the idea of localization, along with a discussion on the connection between our localized GPs and local kernel ridge regression.
The experimental campaign in Section 3 reports results on a variety of benchmarks for regression, and Section 4 concludes the paper.

\section{Gaussian Processes, Kernel Ridge Regression, and Localization}
\label{main}

\subsection{Gaussian Process Regression}
A zero-mean Gaussian process (GP) ($f(\bx):\bx\in \mathcal{X} \subset \mathbb{R}^d$) is a set of random variables $f(\bx)$ indexed by the input set $\mathcal{X}$ such that for each finite subset $\{\bx_1, \ldots, \bx_m\} \subset \mathcal{X}$ the random vector $(f(\bx_1),\dots,f(\bx_m))$ is a zero-mean multivariate normal. The finite-dimensional distribution of such a process is determined by the covariance function $K:\mathcal{X}\times\mathcal{X} \rightarrow \mathbb{R},$ defined by
\begin{equation}
    K(\bx,\bx')=E[f(\bx)f(\bx')].
\end{equation}
The fact that $f(\bx)$ is Gaussian process with covariance kernel $K$ is commonly denoted by 
\[
f(\bx) \sim \mathcal{GP}(0,K(\bx,\bx')).
\]

Given a data set $\mathcal{D}$ comprising a set of input-label pairs $\mathcal{D} = \{(\bx_i, y_i)\}_{i = 1,\ldots,n}$, GPs can be used as a prior over functions to model the relationship between inputs and labels. 
The likelihood function for GPs applied to regression tasks can be derived from assuming that
\begin{equation}\label{GP_model}
y_i=f(\bx_i)+\varepsilon_i, \qquad \varepsilon_i \sim \mathcal{N}(0, \sigma^2).
\end{equation}
The GP prior, together with the Gaussian assumption on the relationship between $f(\bx)$ and $y_i$ induces a multivariate normal distribution over $\by = (y_1, \ldots, y_n)^{\top}$ as
\begin{equation}
    \by \sim \mathcal{N}(0, K_{XX} + \sigma^2 I),
\end{equation}
where $(K_{XX})_{ij} = K(\bx_i, \bx_j)$.

In GP regression it is possible to compute the predictive
distribution of the function values at any arbitrary input points $\bx \in \mathcal{X}$ given $\mathcal{D}$ by means of well-known formulas for conditional distributions of Gaussian random vectors. 
Given the data set $\mathcal{D}$, it is straightforward to show \cite{Rasmussen06} that the posterior over the function is also a GP
\[
f|\by, X \sim \mathcal{GP}(m(\bx),\mathcal{K}(\bx,\bx'))
\]
with mean
\begin{equation} \label{GP_mean}
    m(\bx)=K_{\bx X}\left(K_{XX}+\sigma^{2} I_{n}\right)^{-1}\mathbf{y}, \quad \bx\in \mathcal{X},
\end{equation}
and covariance function
\begin{equation} \label{GP_kernel}
\mathcal{K}\left(\bx, \bx^{\prime}\right) =K\left(\bx, \bx^{\prime}\right)-K_{\bx X}\left(K_{XX}+\sigma^{2} I_{n}\right)^{-1} K_{X \bx^{\prime}}, \quad \bx, \bx^{\prime} \in \mathcal{X},
\end{equation}
where $K_{X \bx}=K_{\bx X}^{\top}=\left(K\left(\bx_{1}, \bx\right), \ldots, K\left(\bx_{n}, \bx\right)\right)^{\top}.$ 

The problem with the expressions above is that they require solving a linear system involving a matrix of size $n \times n$. 
Direct methods to solve these operations require $O\left(n^{3}\right)$ operations and storing $O\left(n^{2}\right)$. 
Iterative solvers, instead, can reduce these complexities by relying exclusively on matrix-vector products, which require $O\left(n^{2}\right)$ operation per iteration and do not need to store the Gram matrix \cite{FilipponeICML15,Cutajar16}.
However, a quadratic time complexity may still prohibitive for large-scale problems.

There is a rich literature on approaches that recover tractability by introducing approximations. 
One popular line of work introduces $m$ so-called inducing-points as a means to approximate the whole GP prior \cite{quinonero2005unifying}. 
This treatment of GPs was later extended within a scalable variational framework \cite{Titsias09,Krauth17}, making the complexity cubic in the number of inducing points $m$.
Another approach, proposes ways to linearize GPs by obtaining an explicit set of features so as to obtain a close approximation to the original kernel-based model. 
Within this framework, a popular approach is based on random features \cite{Rahimi08}. 
Denoting by $\Phi$ the $n \times D$ matrix obtained by applying a set of $D$ random basis functions to the inputs in $\bx_1 \ldots, \bx_n$, these approximations are so that $\Phi \Phi^{\top}$ approximates in an unbiased way the original kernel matrix $K_{XX}$, that is $E [\Phi \Phi^{\top}] = K_{XX}$.
For the Gaussian kernel, for example, a Fourier analysis shows that the basis functions that satisfy this property are trigonometric functions with random frequencies \cite{Rahimi08}. 
This approach has been applied to GPs in \cite{Gredilla10} and later made scalable by operating on mini-batches in \cite{Cutajar17}.

\subsection{Locally Smoothed Gaussian Process Regression}
GP regression as formulated above is an example of a \textit{global learner}; for a fixed training data $\mathcal{D}$ it builds a posterior distribution over functions that can be used to calculate the predictive distribution for any test inputs. 
To construct a predictive distribution for \textit{any} input point it is necessary to use all the information available from the data $\mathcal{D}$, resulting in the need to do algebraic operations with the matrix $\left(K_{XX}+\sigma^{2} I_{n}\right)$. 
However, if we focus on the prediction problem locally, at a \textit{given} target input $\bx_0$, most of the information carried by the (potentially large) covariance matrix might be neglected with little loss of information. 
The main idea behind localized GPs is to down-weight the contribution of the  data points far from $\bx_0,$ so that the structure of the covariance matrix is more adapted to the prediction task at a given point. To make this general idea work, we need to tackle two challenges. First, we need to specify what it means to be far or close to a given point; second, the change of the structure of the covariance matrix must give us a valid covariance matrix, i.e., the resulting covariance matrix should be symmetric and positive definite.
Note that a simple truncation of the covariance function to obtain a compact-support covariance function may generally destroy positive definiteness \cite{Kaufman08}.  

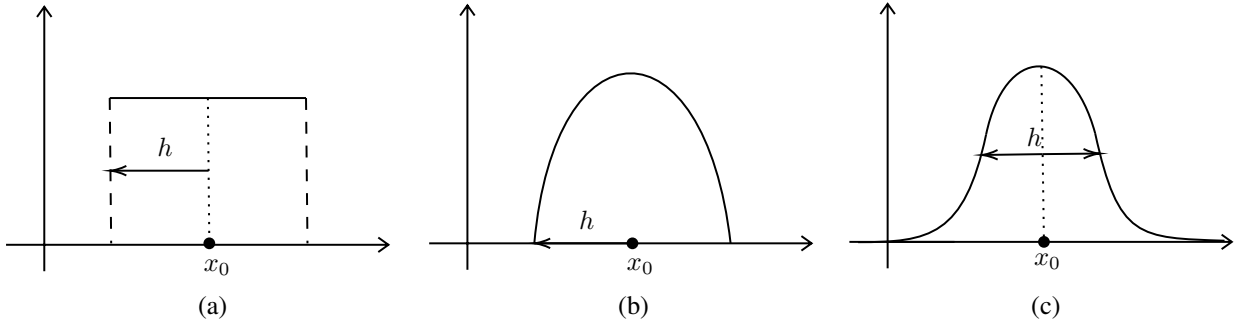
\begin{figure}
\tikzset{every picture/.style={line width=0.75pt}} 

\begin{tikzpicture}[x=0.55pt,y=0.55pt,yscale=-1,xscale=1]

\draw  (8,235.8) -- (271,235.8)(34.3,72) -- (34.3,254) (264,230.8) -- (271,235.8) -- (264,240.8) (29.3,79) -- (34.3,72) -- (39.3,79)  ;
\draw    (79.3,134.8) -- (214.3,134.8) ;
\draw  [dash pattern={on 4.5pt off 4.5pt}]  (79.3,134.8) -- (80.3,234.8) ;
\draw  [dash pattern={on 4.5pt off 4.5pt}]  (214.3,134.8) -- (215.3,234.8) ;
\draw  (299,234.8) -- (562,234.8)(325.3,71) -- (325.3,253) (555,229.8) -- (562,234.8) -- (555,239.8) (320.3,78) -- (325.3,71) -- (330.3,78)  ;
\draw    (371.3,234.8) .. controls (385.3,79.8) and (488.3,77.8) .. (506.3,234.8) ;
\draw  (588,233.8) -- (851,233.8)(614.3,70) -- (614.3,252) (844,228.8) -- (851,233.8) -- (844,238.8) (609.3,77) -- (614.3,70) -- (619.3,77)  ;
\draw    (614.3,233.8) -- (660.3,233.8) ;
\draw    (682,160) .. controls (695,98) and (740,97) .. (757,159) ;
\draw    (682,160) .. controls (667,230) and (643,234) .. (594,234) ;
\draw    (757,159) .. controls (773,232) and (787,231) .. (846,233) ;
\draw    (147.8,234.8) ;
\draw [shift={(147.8,234.8)}, rotate = 0] [color={rgb, 255:red, 0; green, 0; blue, 0 }  ][fill={rgb, 255:red, 0; green, 0; blue, 0 }  ][line width=0.75]      (0, 0) circle [x radius= 3.35, y radius= 3.35]   ;
\draw    (438.8,234.8) ;
\draw [shift={(438.8,234.8)}, rotate = 0] [color={rgb, 255:red, 0; green, 0; blue, 0 }  ][fill={rgb, 255:red, 0; green, 0; blue, 0 }  ][line width=0.75]      (0, 0) circle [x radius= 3.35, y radius= 3.35]   ;
\draw    (721.8,233.8) ;
\draw [shift={(721.8,233.8)}, rotate = 0] [color={rgb, 255:red, 0; green, 0; blue, 0 }  ][fill={rgb, 255:red, 0; green, 0; blue, 0 }  ][line width=0.75]      (0, 0) circle [x radius= 3.35, y radius= 3.35]   ;
\draw  [dash pattern={on 0.84pt off 2.51pt}]  (146.8,134.8) -- (147.8,234.8) ;
\draw    (147.3,184.8) -- (81.8,184.8) ;
\draw [shift={(79.8,184.8)}, rotate = 360] [color={rgb, 255:red, 0; green, 0; blue, 0 }  ][line width=0.75]    (10.93,-3.29) .. controls (6.95,-1.4) and (3.31,-0.3) .. (0,0) .. controls (3.31,0.3) and (6.95,1.4) .. (10.93,3.29)   ;
\draw    (438.8,234.8) -- (373.3,234.8) ;
\draw [shift={(371.3,234.8)}, rotate = 360] [color={rgb, 255:red, 0; green, 0; blue, 0 }  ][line width=0.75]    (10.93,-3.29) .. controls (6.95,-1.4) and (3.31,-0.3) .. (0,0) .. controls (3.31,0.3) and (6.95,1.4) .. (10.93,3.29)   ;
\draw  [dash pattern={on 0.84pt off 2.51pt}]  (720,113) -- (721.8,233.8) ;
\draw    (720.9,173.4) -- (681,173.97) ;
\draw [shift={(679,174)}, rotate = 359.18] [color={rgb, 255:red, 0; green, 0; blue, 0 }  ][line width=0.75]    (10.93,-3.29) .. controls (6.95,-1.4) and (3.31,-0.3) .. (0,0) .. controls (3.31,0.3) and (6.95,1.4) .. (10.93,3.29)   ;
\draw    (720.9,173.4) -- (758,173.02) ;
\draw [shift={(760,173)}, rotate = 179.41] [color={rgb, 255:red, 0; green, 0; blue, 0 }  ][line width=0.75]    (10.93,-3.29) .. controls (6.95,-1.4) and (3.31,-0.3) .. (0,0) .. controls (3.31,0.3) and (6.95,1.4) .. (10.93,3.29)   ;

\draw (142,244) node [anchor=north west][inner sep=0.75pt]   [align=left] {$\displaystyle x_{0}$};
\draw (713,240) node [anchor=north west][inner sep=0.75pt]   [align=left] {$\displaystyle x_{0}$};
\draw (433,243) node [anchor=north west][inner sep=0.75pt]   [align=left] {$\displaystyle x_{0}$};
\draw (110,160) node [anchor=north west][inner sep=0.75pt]   [align=left] {$\displaystyle h$};
\draw (400,210) node [anchor=north west][inner sep=0.75pt]   [align=left] {$\displaystyle h$};
\draw (708,153) node [anchor=north west][inner sep=0.75pt]   [align=left] {$\displaystyle h$};
\draw (138,269) node [anchor=north west][inner sep=0.75pt]   [align=left] {(a)};
\draw (427,269) node [anchor=north west][inner sep=0.75pt]   [align=left] {(b)};
\draw (711,269) node [anchor=north west][inner sep=0.75pt]   [align=left] {(c)};

\end{tikzpicture}
\caption{Examples of local kernels: (a) Rectangular kernel $k(\bx)=\mathbb{I}(\|\bx\| \leq 1).$ (b) Epanechnikov kernel $k(\bx) = \frac{d+2}{2V_d}(1 - \|\bx\|^2)  \mathbb{I}{(\|\bx\| \leq 1)}.$ (c) Gaussian kernel $k(\bx) = \frac{1}{2\pi}e^{-\|\bx\|^2}.$}
\end{figure}

We accomplish localization of GPs in a straightforward manner as follows. We localize the target and the prior in the model (\ref{GP_model}) by multiplying them by the square root of the weighting function 
\[
k_h(\bx,\bx_0):=\frac{1}{h}k\left(\frac{\|\bx-\bx_0\|}{h} \right),
\]
where $k:\mathcal{X} \subset \mathbb{R}^d \rightarrow \mathbb{R} $ is a non-negative, integrable function satisfying $\int K(\bx)d\bx = 1$ and $\|\cdot\|$ is Euclidean norm on $\mathbb{R}^d.$ 
Considering the square root of the weighting function will be convenient later when we discuss the link between local GPs and local Kernel Ridge Regression. 
Some classical examples of the weighting functions are given in Fig.~1. 
Because of the linearity of the weighting operation, the resulting model is another zero-mean Gaussian process $\tilde{f}(\bx)$ with covariance function given by
\begin{equation}\label{loc_ker}
    \tilde{K}(\bx,\bx';\bx_0)=k_h^{\frac{1}{2}}(\bx,\bx_0)K(\bx,\bx')k_h^{\frac{1}{2}}(\bx',\bx_0).
\end{equation}

In this formulation, we have localized the relationship between noisy targets and function realizations as
\begin{equation}\label{local_model}
    \tilde{y}_i=\tilde{f}(\bx_i)+\varepsilon_i, \quad \varepsilon_i \sim \mathcal{N}( 0,\sigma^2),
\end{equation}
with $\tilde{y}_i=\sqrt{k_h(\bx_i,\bx_0)}y_i$, and the prior is given by a zero-mean GP with the localized covariance kernel (\ref{loc_ker}). The model (\ref{local_model}) can be alternatively written as a model with heteroscedastic noise 
\[
    y_i=f(\bx_i)+\frac{1}{\sqrt{k_h(\bx_i,\bx_0)}}\varepsilon_i, \quad \varepsilon_i \sim \mathcal{N}( 0,\sigma^2).
\]
Making the noise parameter location-dependent can significantly improve the performance for problems where the assumption of a homoscedastic noise is not satisfied.
\begin{figure}[!t]
    \centering
    \includegraphics[width=16.5cm]{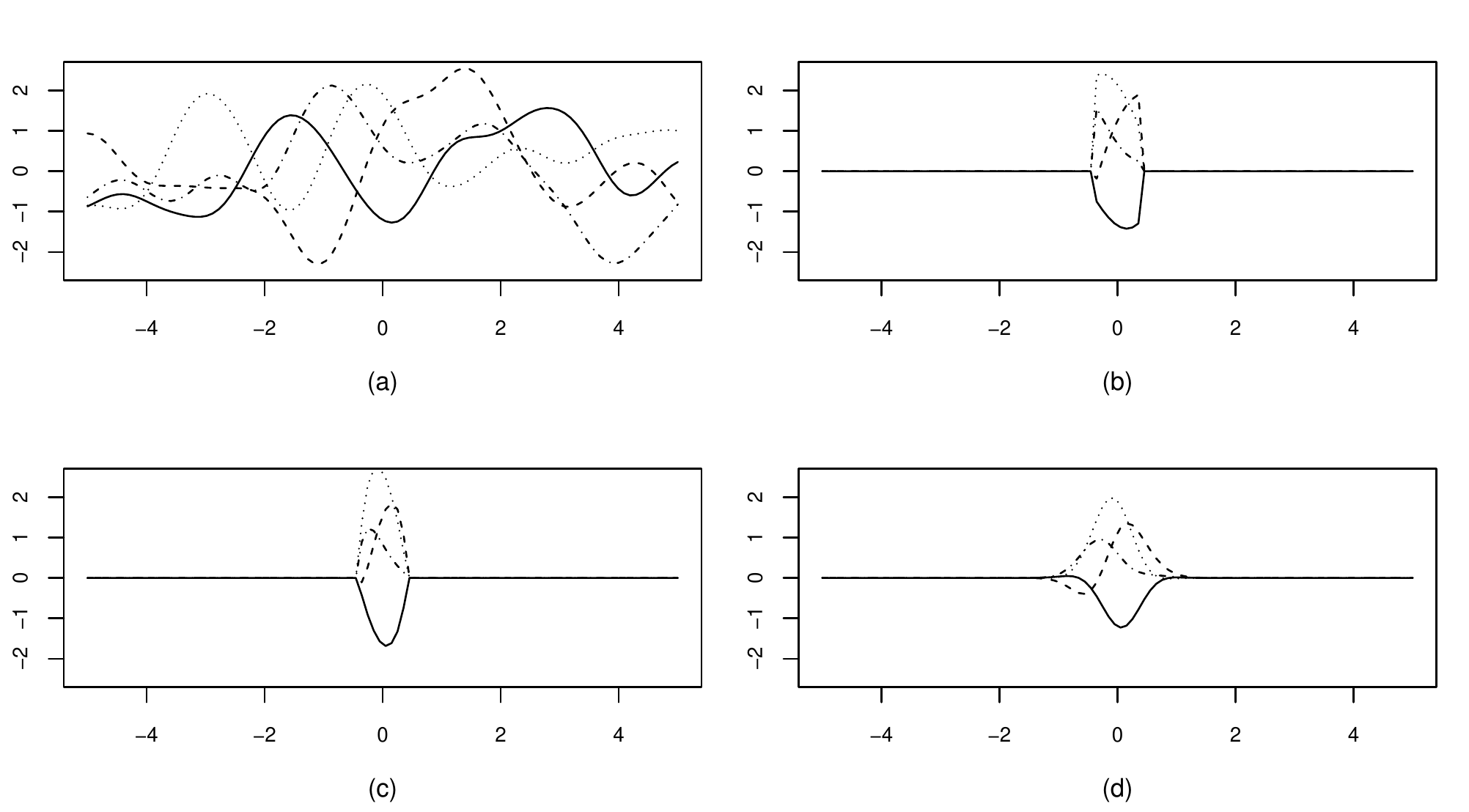}
    \caption{(a) Samples from a global GP prior with exponential kernel. (b) Samples from a GP prior with exponential kernel localized by rectangular smoother centered at $x_0=0.$ (c) Samples from a GP prior with exponential kernel localized by Epanechnikov smoother centered at $x_0=0.$ (d) Samples from a GP prior with exponential kernel localized by Gaussian smoother centered at $x_0=0.$}
    \label{fig:my_label}
\end{figure}

\begin{theorem}
Let $I = \{i:\|\bx_i-\bx_0\| \leq h\},$  $X_I = \{\bx_i:i \in I\}$ and $\mathbf{y}_I = \{y_i\}_{i\in I} \in \mathbb{R}^{|I|}.$ Assume that (\ref{local_model}) holds for the fixed target point $\mathbf{x}_0.$ Then 
$
f(\mathbf{x}_0) \mid \mathbf{y}_I
$
is a Gaussian random variable with mean and variance given by
\begin{eqnarray}
     \tilde{m}(\bx_0)&=&K_{\bx_0 X_I}\left(K_{X_I X_I}+\sigma^{2} W_{\bx_0}^{-1}\right)^{-1}\mathbf{y}_I
\\
    \tilde{\mathcal{K}}\left(\bx_0, \bx_0\right) &=& K\left(\bx_0, \bx_0\right)-K_{\bx_0 X_I}\left(K_{X_I X_I}+\sigma^{2} W_{\bx_0}^{-1}\right)^{-1} K_{X_I \bx_0}
\end{eqnarray}
where $W_{\bx_0}$ is the diagonal matrix with main diagonal entries $k_h(\bx_i,\bx_0),\,\, \bx_i \in  X_I.$
\end{theorem}
\textbf{Proof:} Let $\bx_{0}$ be any fixed target point. Then the observations $\mathbf{y}_I \in \mathbb{R}^{|I|}$ and GP-function value at target point $f_{0}=f\left(\bx_{0}\right) \in \mathbb{R}$ are jointly Gaussian such that
$$
\left[\begin{array}{c}
\by \\
f_{0}
\end{array}\right] \sim \mathcal{N}\left(\left[\begin{array}{l}
\textbf{0}_I \\
0
\end{array}\right],\left(\begin{array}{cc}
K_{X_I X_I} + \sigma^2 W_{\bx_0}^{-1} & K_{X_I \bx_0}  \\
K_{\bx_0 X_I} &
K(\bx_0,\bx_0)
\end{array}\right)\right).
$$
Then the proposition follows from the basic formula for conditional distributions of Gaussian random vectors (see, e.g., \cite{Rasmussen06}, Appendix A.2).

Compared to global GPs, in the local formulation, in order to compute the posterior mean and variance, we need to invert $\left(K_{X_I X_I}+\sigma^{2} W_{\bx_0}^{-1}\right).$ This might give a key advantage when dealing with large data sets, as the localization by compactly supported kernel (local) could significantly sparsify the Gram matrix corresponding to $K_{XX}$. 
Denoting by $s_0$ the number of inputs for which the localizing weights are nonzero for a test point $\mathbf{x}_0$, the complexity of performing such an inversion is $\mathcal{O}(s_0^3)$. 
Another interesting observation is that the kernel function $\tilde{K}\left(\bx, \bx^{\prime}\right)$ is potentially more flexible than the original kernel function; this is due to the multiplication by the localizing weighting function, which may introduce some interesting nonstationarity even for kernel functions which are stationary, depending on the choice of the weighting function.

The calculation of the predictive distribution in with Locally Smoothed Gaussian Process Regression (LSGPR) is described in Algorithm~\ref{alg:LSGPR}
The parameter selection in probabilistic models given by GPs is based on the {\em marginal log-likelihood} maximization, which, in our local formulation, can be defined as follows
\[
    \log p(\mathbf{y}_I|X_I) = -\frac{1}{2} \mathbf{y}_I^{\top}\left(K_{X_IX_I} + \sigma^2 W_{\mathbf{x}_0}^{-1}\right)^{-1}\mathbf{y}_I - \frac{1}{2}\log\left|K_{X_IX_I} + \sigma^2 W_{\mathbf{x}_0}^{-1}\right| -\frac{n}{2} \log (2 \pi)
\]
Unfortunately, gradient-based optimization cannot be used to find the optimal localization parameter $h$, as the marginal log-likelihood is not continuously differentiable w.r.t. this parameter when compactly supported local kernels are used. 
The simplest way to resolve this problem is by using grid search for the localization parameter $h$, while kernel parameters can be optimized by gradient-based methods for any given $h.$
\begin{algorithm}[H]
            \small
            \caption{LSGPR}\label{alg:LSGPR}
            \begin{algorithmic}[1]
                \State \textbf{Input:} $X$, $\mathbf{y}$, $\sigma^2$, $h$, $\mathbf{x}_0$
                \State \textbf{Output:} $\tilde{m}(\mathbf{x}_0)$, $\tilde{\mathcal{K}}(\mathbf{x}_0, \mathbf{x}_0)$
                \State $I:=\{i:||\mathbf{x}_i - \mathbf{x}_0|| \leq h\}$
                \State $X_I := \{\mathbf{x}_i:i \in I\}$
                \State $\mathbf{y}_I := \{y_i:i \in I\}$
                \State $W_{\bx_0}:=\mathrm{diag}(\{k_h(\mathbf{x}_i,\mathbf{x}_0): i \in I\}) $
                \State L:=Cholesky$\left(K_{X_I X_I}+\sigma^{2} W_{\bx_0}^{-1}\right)$
                \State $\bm{\alpha} := (L^{-1})^{\top} L^{-1} \mathbf{y}$
                \State $\tilde{m}(\mathbf{x}_0) := K_{\mathbf{x}_0, X_I}\bm{\alpha}$
                \State $\mathbf{v}$ := $L^{-1}K_{\mathbf{x}_0, X_I}$
                \State $\tilde{\mathcal{K}}(\mathbf{x}_0, \mathbf{x}_0)$ := $K(\mathbf{x}_0, \mathbf{x}_0) - \mathbf{v}^{\top}\mathbf{v}$
            \end{algorithmic}
\end{algorithm}

\subsection{Local Kernel Ridge Regression}
For every Gaussian process $f(\bx)$ with covariance function $K(\bx,\bx')$ there is a unique corresponding Hilbert space $\mathcal{H}_K$. This is commonly referred to as a \textit{reproducing kernel Hilbert space} (RKHS) and constructed as a completion of the linear space of all functions
\[
\bx \mapsto \sum_{i=1}^{k} \alpha_{i} K\left(\ba_{i}, \bx\right), \quad \alpha_{1}, \ldots, \alpha_{k} \in \mathbb{R}, \ba_{1}, \ldots, \ba_{k} \in \mathcal{X}, k \in \mathbb{N}
\]
relative to the norm induced by the inner product
\[
\left\langle\sum_{i=1}^{k} \alpha_{i} K\left(\bs_{i}, \cdot\right), \sum_{j=1}^{l} \beta_{j} K\left(\bt_{j}, \cdot\right)\right\rangle_{\mathcal{H}_K}=\sum_{i=1}^{k} \sum_{j=1}^{l} \alpha_{i} \beta_{j} K\left(\bs_{i}, \bt_{j}\right).
\]

It is well know that the posterior mean of Gaussian process regression can be alternatively derived by minimizing the regularized empirical risk over the RKHS \cite{kimeldorf1970correspondence}; see, e.g., \cite{kanagawa2018gaussian} for a recent review. For local GPs, this corresponds to a weighted least square minimization over the RKHS with the weights given by $k_h(\bx,\bx_0)$, that is 
\begin{equation}\label{loc_risk_min}
    \tilde{m}(\bx)=\arg \min_{f \in \mathcal{H}_k}\sum_{i=1}^n \left(y_i-f(\bx_i)\right)^2k_h\left(\bx_i,\bx_0\right)+\frac{\sigma^2}{n} \|f\|_{\mathcal{H}_k}^2.
\end{equation}

Note that in the local formulation, for a given point $\bx_0$ one has to estimate both the parameters of the reproducing kernel and the width of the local kernel $h.$
Here are two examples of well-known classical local methods which are the solution of the empirical risk minimization problem (\ref{loc_risk_min}).

\paragraph{K-nearest neighbors}
This model corresponds to the noise-free case ($\sigma = 0$) with a positive constant reproducing kernel and a rectangular local kernel whose width is adjusted to contain exactly $k$ data points. The solution of the minimization problem (\ref{loc_risk_min}) is the mean of the outputs corresponding to the $k$ closest to $\bx_0$ input points.

\paragraph{Local polynomial regression} 
If we use the polynomial kernel $K(\bx,\bx') = \left(1+\bx \bx^{\prime}\right)^{k}$ for the space $\mathcal{H}_K,$ and use any smooth local kernel (i.e. exponential), then in the noise-free case the solution of the minimization problem (\ref{loc_risk_min}) is so called local polynomial regression \cite{Tsybakov09}. In this special case, when the degree of the polynomial is $0,$ we have Nadaraya-Watson regression, which is the minimizer of the local squared loss over the constant function.

\section{Experiments}
    \subsection{Toy dataset}

        \begin{figure}[t]
        \centerline{\includegraphics[width=0.9\textwidth]{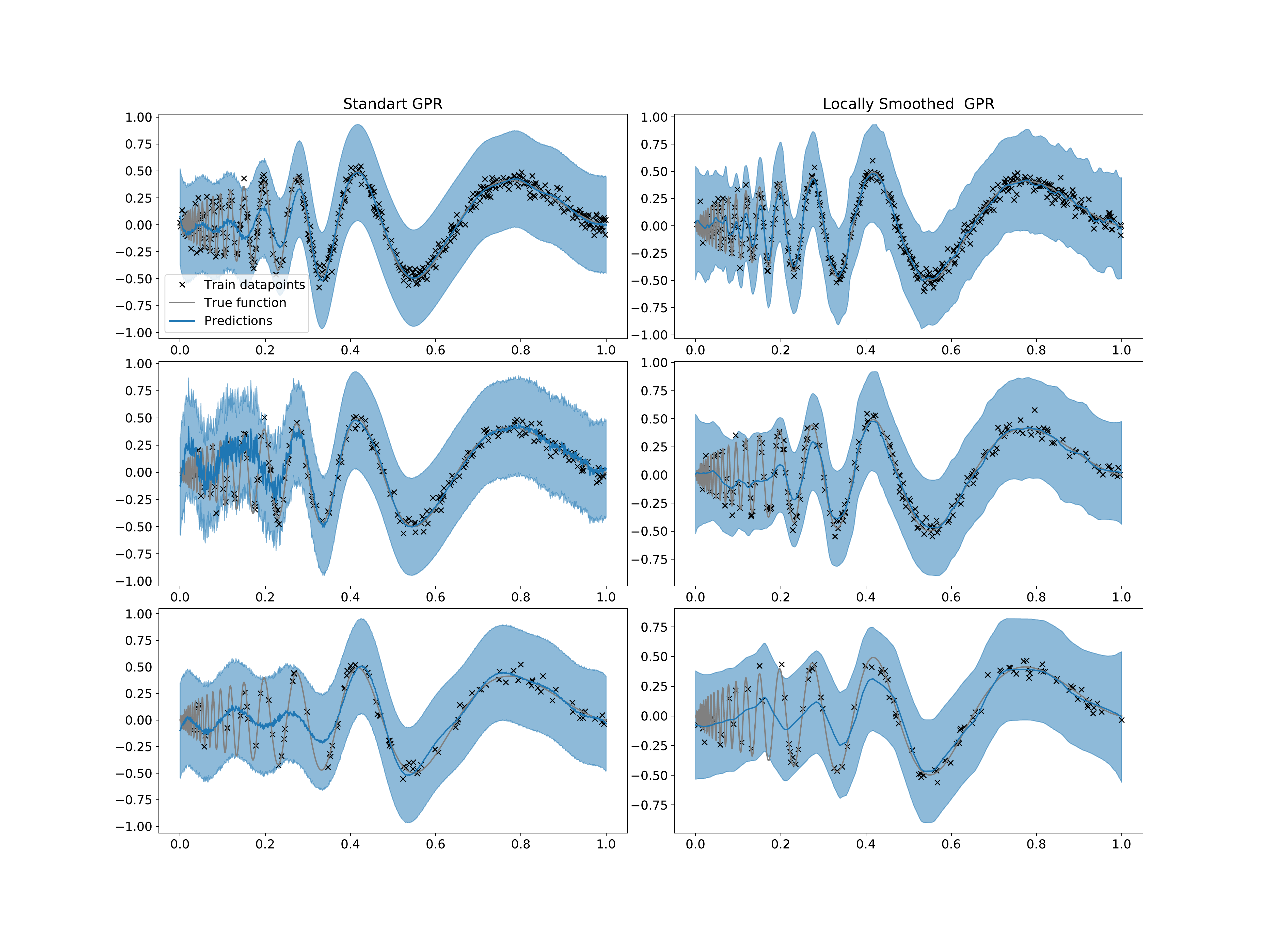}}
            \caption{Illustration of the predictive distribution of GPs (left) and LSGPs (right) applied to data sampled from the Doppler function for 400 training points (top), 200 training points (middle) and 100 training points. }
            \label{fig:toy}
        \end{figure}

        In order to illustrate the behavior of the proposed Locally Smoothed Gaussian Process (LSGPs), we start from a toy dataset generated from the Doppler function (Fig. \ref{fig:toy}).
        \begin{equation}
            y(x)=\sqrt{x(1-x)}\sin\left(\frac{2.1\pi}{x+0.05} \right) + \varepsilon, \quad 0 \leq x \leq 1, \quad \varepsilon \sim \mathcal{N}(0, 0.1)
        \end{equation}
        For this experiment, we used the RBF kernel and the Epanechnikov localizing kernel 
        \begin{equation}
            k(x) = \frac{3}{4}(1 - |x|^2)  \mathbb{I}{(|x| \leq 1)}. 
        \end{equation}
        We tuned the lengthscale parameter of the RBF kernel by optimizing the marginal log-likelihood of the model. We used the L-BFGS algorithm for gradient optimization \cite{pytlak2008conjugate}. We chose the value of the parameter $h$ of the localizing kernel that gave the best Mean Squared Error (MSE) on a validation set. The LSGP model used on average 7 training points to make a prediction. We compared the predictions of LSGP with the predictions of standard GP regression.

        As we can see from Fig. (\ref{fig:toy}), the Gaussian Process with RBF kernel is unable to make reasonable predictions in the region where the target function contains high frequency components.
        While some nonstationary covariance functions might be appropriate for this example, the combination of a standard stationary covariance function with the localization approach offers substantial modeling improvements.  
        
    \subsection{UCI datasets}
        We evaluated the performance of the LSGP method on several problems from the UCI datasets collection and compared it against standard GPR, Deep GPs approximated with random features \cite{Cutajar17}, and k-nearest neighbors (KNN) regression. 
        In particular, we aim to compare the predictive performance offered by the localization against the baseline of exact GPR, and to verify that any performance gains are not just due to localization, meaning that we expect to outperform KNN.  Because our model is more flexible than standard GPR, we also added to the comparison Deep GP models based on random features expansion. 
        The size and dimensionality of these problems is outlined in Table~\ref{tab: UCIdatasets}. Since the Euclidean norm used in the local kernels depends on the units in each coordinate, all the datasets except the Protein were scaled within the $[0,1]$ range. We used a standardization procedure for the Protein dataset because the baseline model worked much better with this type of preprocessing.
        For the Deep GP model we used the hyperparameters and the data preprocessing described in the original paper.
        
        The LSGPR method requires creating a new local model with its own set of hyperparameters for each input point where the prediction has to be made. During the optimization, the kernel parameters of each model were constrained to be equal among all local models. Considering the hyperparameter $h$, we found that it is hard to find values of $h$ which perform well across all regions of the input space. Thus, for each input point of interest, we chose values of $h$ that ensured that the localizer considers at least $m$ neighboring training points. In this experiment, we used 3-fold cross-validation to choose the noise variance $\sigma^2$, the lengthscale of the GP kernel and the parameter $m$ of the localizing kernel. We report the results on the held-out test set.  
        
        In this experiment, we also used the Hilbert localizing kernel \cite{shepard1968two,devroye1998hilbert}
        \begin{equation}
        k(\bx) = \| \bx \|^{-1}   \mathbb{I} {(\| \bx \| \leq 1)},
        \end{equation}
        which showed good performance for most of the datasets. In Table \ref{tab:UCImse} locally smoothed Gaussian Process Regression based on Hilbert kernel is referred to as LSGPR Hilbert, while the same model based on Epanechnikov kernel is referred as LSGPR Epanechnikov. GP regression, Deep GP regression and KNN regression are referred to as GP, DeepGP and KNN, respectively.
        
        The results indicate that LSGP offers competitive performance with respect to GPR and Deep GP baselines. 
        The results also clearly show that LSGP offers superior performance to KNN, suggesting that the localization alone is not enough to obtain good performance, and that this works well in combination with the GP model. 
        To make a comparison with the baselines, we used the one-sided Wilcoxon test \cite{10.2307/3001968}. For each method, we measured its performance on 10 data splits, and we used exactly the same splits for testing performance of each method, so we had matched samples of the MSEs.
        Then we used the test to compare methods in pairs, where the alternative hypothesis was that the MSE of any given method is smaller than the MSE of a competitors.
        We used confidence level $\alpha=0.05$. In Table \ref{tab:UCImse} the results that are statistically better than the competitors are marked in bold.
        \begin{table}[t]
        \caption{UCI datasets used for evaluation.}
        \begin{tabular*}{\hsize}{@{\extracolsep{\fill}}lll@{}}
            \toprule
            Dataset &  Training instances &  Dimensionality \\
            \hline
            Yacht & 308 & 6  \\
            Boston & 506 & 13   \\
            Concrete  & 1030 & 8 \\
            Kin8nm  & 8192 & 8 \\
            Powerplant  & 9568 & 4 \\
            Protein  &  45730 & 9 \\
            \hline
        \end{tabular*}
        \label{tab: UCIdatasets}
        \end{table}
        
        \begin{table}[t]
            \caption{Comparison in terms of test set MSE between LSGPR a standard GPR.}
            \begin{tabular*}{\hsize}{@{\extracolsep{\fill}}llllll@{}}
            \toprule
            Dataset &  LSGPR Hilbert & LSGPR Epanechnikov & GP & DeepGP& KNN \\
            \hline
            Yacht &  \textbf{0.63$\pm$0.12} & 2.02$\pm$0.58 & 1.09$\pm$0.05 & 0.93$\pm$0.13 & 57.80$\pm$16.65\\
            Boston &  14.78$\pm$0.88 & 15.30$\pm$1.28& 17.94$\pm$0.71 & \textbf{7.92$\pm$0.14} & 23.30$\pm$2.58  \\
            Concrete  & \textbf{34.79$\pm$1.07} & 40.43$\pm$3.16& 37.81$\pm$0.61 & 130.94$\pm$3.93 & 94.23$\pm$7.89 \\
            Kin8nm  & 0.01$\pm$0.000 & 0.01$\pm$0.00 & 0.01$\pm$0.000 & 0.06$\pm$0.00 & 0.01$\pm$0.00 \\
            Powerplant  & 14.65$\pm$0.34  & \textbf{14.40$\pm$0.67}& 16.85$\pm$0.69 & 14.57$\pm$0.15 & 15.35$\pm$0.49 \\
            Protein  & 36.85$\pm$2.15 & \textbf{12.50$\pm$0.26} & 17.03$\pm$0.57 & 16.94 $\pm $0.16 & 19.89$\pm$4.26\\
            \hline
            \end{tabular*}
            \label{tab:UCImse}
        \end{table}
    
    \section{Conclusions}
    
    In this work we developed a novel framework to localize Gaussian processes (GPs). 
    We focused in particular on Gaussian Process Regression (GPR), and we derived the GP model after applying the localization operation through the down-weighting of contributions from input points which are far away from a given test point. 
    The form of the localized GP maintains positive definiteness of the covariance, and it allows for considerable speedups compared to standard global GPR due to the sparsification effect of the Gram matrix. 
    
    The proposed method requires cross-validation to tune the scale parameter of localizing kernel, while others GP-based techniques use a less expensive marginal log-likelihood (MLL) gradient optimization to tune these types of parameters. We found MLL gradient optimization problematic because of the discontinuity of local kernel with respect to the scale parameter, which in turn makes MLL function discontinuous with respect to this parameter.
    It would be interesting to investigate ways to extend the idea of localization for GPR to other tasks, such as classification.


\bibliography{sample,filippone}
\bibliographystyle{unsrt}

\end{document}